\documentclass[a4paper,conference]{IEEEtran}
\IEEEoverridecommandlockouts

\ifdefined\UseShortAcronyms

\else

\fi
\newcommand{\ie}{{\em i.e.}}
\newcommand{\eg}{{\em e.g.}}

\usepackage{color}
\definecolor{red}{rgb}{1.00,0.20,0.20}
\definecolor{blue}{rgb}{0.20,0.20,1.00}
\definecolor{green}{rgb}{0.00,1.00,0.00}

\usepackage{array}
\newcolumntype{L}[1]{>{\raggedright\let\newline\\\arraybackslash\hspace{0pt}}m{#1}}
\newcolumntype{C}[1]{>{\centering\let\newline\\\arraybackslash\hspace{0pt}}m{#1}}
\newcolumntype{R}[1]{>{\raggedleft\let\newline\\\arraybackslash\hspace{0pt}}m{#1}}

\usepackage{amsmath,amsfonts,bm,amssymb,xspace}

\def\secref#1{section~\ref{#1}}

\def\eqref#1{equation~\ref{#1}}

\def\1{\bm{1}}

\def\rvs{{\mathbf{s}}}

\def\rvx{{\mathbf{x}}}
\def\rvy{{\mathbf{y}}}

\DeclareMathAlphabet{\mathsfit}{\encodingdefault}{\sfdefault}{m}{sl}
\SetMathAlphabet{\mathsfit}{bold}{\encodingdefault}{\sfdefault}{bx}{n}

\newif\ifdraft
\ifdraft
    \usepackage[colorinlistoftodos]{todonotes}
    \newcommand{\todoline}[1]{\todo[inline]{#1}}
\else
    \newcommand{\todoline}[1]{}
\fi

\usepackage{cite}
\usepackage{url}
\usepackage{comment}
\usepackage{graphicx}
\usepackage{amsmath,amssymb,amsfonts}
\usepackage{float}
\usepackage{booktabs}
\usepackage{dsfont}
\usepackage{multirow}

\newcommand{\hitrate}[1]{HitRate\textsubscript{#1, 2m}\xspace}
\newcommand{\ade}[1]{minADE\textsubscript{#1}\xspace}

\ifCLASSINFOpdf
\else
\fi

\begin{document}

\title{Trajectory Prediction for Autonomous Driving with Topometric Map}

\author{
	\IEEEauthorblockN{Jiaolong Xu,
	Liang Xiao,
	Dawei Zhao,
	Yiming Nie and
	Bin Dai}
}


\maketitle

\begin{abstract}
State-of-the-art autonomous driving systems rely on high definition (HD) maps for localization and navigation. However, building and maintaining HD maps is time-consuming and expensive. Furthermore, the HD maps assume structured environment such as the existence of major road and lanes, which are not present in rural areas. In this work, we propose an end-to-end transformer networks based approach for map-less autonomous driving. The proposed model takes raw LiDAR data and noisy topometric map as input and produces precise local trajectory for navigation. We demonstrate the effectiveness of our method in real-world driving data, including both urban and rural areas. The experimental results show that the proposed method outperforms state-of-the-art multimodal methods and is robust to the perturbations of the topometric map. The code of the proposed method is publicly available at \url{https://github.com/Jiaolong/trajectory-prediction}.
\end{abstract}

%
\IEEEpeerreviewmaketitle

\section{Introduction}
Autonomous driving is one of the most impactful technique that is going to improve our lives. The vast majority of industry-led autonomous driving techniques are based on building and maintaining high definition (HD) maps. Most of them assume a structured environment, {\eg} the existence of major roads and lanes. However, there is a significant portion of the road network that cannot satisfy such requirement, such as the rural areas. On the other hand, building and maintaining HD maps is time-consuming and expensive, especially for the environments with rapid rate of changes.

There have been many attempts to address this problem. For example, the neural network-based driving approaches \cite{conditional_imitation:2019, mit_variational:2019, learn_drive:2019} take the sensor data ({\eg}, LiDAR, images) as input and outputs control signals, {\eg} steering command and acceleration. These methods do not require detailed map, but they lack explainability, verifiable robustness, and require massive amount of training data to generalize. Other approaches focus on motion planning with topometric map. OpenStreetMap (OSM) \cite{Openstreetmap} is widely used in these methods due to its public availability and rich information. For example, in \cite{2010Autonomous}, the building information from OSM is incorporated for autonomous navigation. In \cite{mit_nav_rural:2018} and \cite{maplite}, OSM is combined with sensor-based perception system for local navigation, where the performance relies on accurate road segmentation. 

In this work, we propose an end-to-end deep learning solution for autonomous navigation, which only requires raw sensor data and weak topological prior map. The proposed method can be applied in both urban and rural areas. The rural areas in this work mainly refer to the unstructured environment with unpaved road. Although OSM can be directly used in our method, as proof of concept, we provide a simpler strategy to create the topometric map by manually labelling roads on satellite images. Such manually labelled topometric map is far less accurate than OSM and contains no rich information other than the position of each clicked waypoint. It can be very noisy and has large variances. In this work, we demonstrate that our method can work with such noisy topometric maps and achieve accurate trajectory prediction for local navigation.

\begin{figure*}
    \centering
    \begin{minipage}[!t]{0.8\textwidth}
        \includegraphics[width=\textwidth]{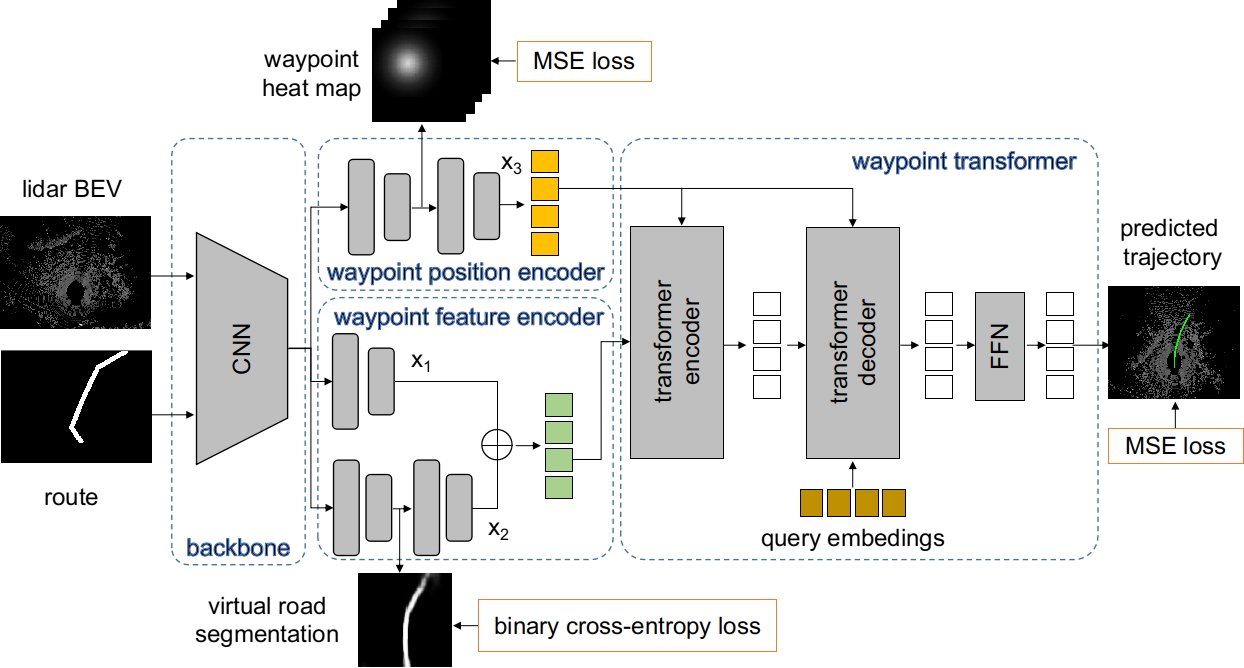}
        \caption{Overview of the proposed transformer-based trajectory prediction model.}
        \label{fig:architecture}
    \end{minipage}
\end{figure*}

Our proposed model is illustrated in Fig.~\ref{fig:architecture}, which is built upon the Transformer encoder-decoder architecture \cite{Transformer}. The input of our model is the raw lidar point clouds and a local route extracted from the noisy topometric map. The output of our model is the future trajectory that the vehicle can follow. We use convolutional neural networks as our backbone to extract features, followed by a transformer module that predicts the final trajectories. 

Two novel auxiliary learning tasks are proposed in our model, including a self-supervised virtual road segmentation task and waypoint heatmap prediction task. The auxiliary learning tasks can significantly boost the trajectory prediction accuracy. The proposed model is end-to-end trainable. Moreover, it does not require manually labelling the ground truth. The required training labels are automatically extracted from recorded driving trajectories, which can save expensive and tedious manual work. In summary, the key contributions of this paper are as follows:

\begin{itemize}
\item State-of-the-art trajectory prediction methods rely on HD maps \cite{2018Multimodal, 2019MultiPath, 2019CoverNet}, we propose an end-to-end trajectory prediction method which only requires coarse topometric maps.
\item As proof of concept, we propose a simple method to create topometric map from satellite images and we demonstrate that our trajectory prediction model is robust to such noisily labelled topometric map.
\item We propose a novel transformer based trajectory prediction model, which outperforms state-of-the-art multimodal methods.
\end{itemize}

\section{Related work}
Our work is related to map-less autonomous driving and learning-based trajectory prediction. 

\subsection{Autonomous driving without HD map}
In \cite{2017Global}, OSM is used for global outer-urban navigation. The performance of this method depends on the quality and correctness of the terrain classification. In \cite{mit_nav_rural:2018}, a map-less driving framework is proposed which combines global sparse topological map with sensor-based perception system for local navigation. A followup work \cite{maplite} utilizes topological map and use road segmentation to register the topometric map in the vehicle frame thus enable local navigation. Other map-less autonomous driving techniques take end-to-end learning approaches \cite{conditional_imitation:2019, mit_variational:2019, learn_drive:2019}. However, these neural network-based driving approaches suffer from the compounding errors, lacking explainability and requiring a high amount of training data to generalize \cite{2019End}.

\subsection{Deep learning-based trajectory prediction}

The success of deep learning in many real-life applications prompts research on trajectory prediction. In \cite{lstm_traj}, the Long Short-Term Memory (LSTM) was successfully applied to predict vehicle locations using past trajectory data. Recently the multimodal approaches \cite{2019MultiPath, 2018Multimodal,2019CoverNet} are becoming popular. However, multimodal regression methods can easily suffer from {\em mode collapse} to a single mode. In \cite{2019CoverNet}, it proposed to predict trajectories by classifying on pre-generated trajectory set. In \cite{2019MultiPath}, the issue is address by using a fixed set of anchor trajectories. In this work, we propose a novel transformer based trajectory prediction model which obtains higher accuracies than state-of-the-art multimodal methods. Most of existing trajectory prediction papers focus on urban driving and rely on rich context information from HD maps. On the contrary, our method is general to both urban and rural environments and does not require HD maps.
\section{Proposed Method}
Our method takes as input the observed raw sensor data and a coarse route extracted from topometric map and predicts a future guidance trajectory that the vehicle can follow. At current stage, our major focus is on the geometric correctness of a predicted future path, thus we assume a constant velocity vehicle and ignore the speed information to simplify the problem setting. 

In this section, we first introduce the creation of topometric map and then describe the details of the proposed trajectory prediction deep learning model.

\subsection{Topometric map}
\label{sec:map}

The topometric map used in our method is a simple graph-like structured data $M = \{V, E\}$, where each vertex $v_i \in \mathbb{R}^2$ represents a waypoint $(x_i, y_i)$ and each edge $e_i$ represents a road segment. Similar to \cite{maplite}, the waypoints are transformed from latitude and longitude to UTM (The Universal Transverse Mercator) coordinates. 

For urban area, it is possible to download the topometric map from OpenStreetMap \cite{Openstreetmap} which provides access to the public. In this work, as proof of concept, we propose a simple method to create topometric maps which can be used for both urban and rural areas. Specifically, we use publicly available satellite images from GoogleMap. By {\em clicking} and connecting some key points on the satellite map, {\eg}, road intersections and sharp turns, we can easily build a sparse global graph. Then, we create a dense topometric map by applying linear interpolation on each edge. It is worth mentioning that it is also possible to create such topometric map in automatic or semi-automatic way, {\eg}, by extracting road from satellite images \cite{VecRoad_20CVPR, 2018RoadTracer}. The created topometric map is quite noisy and imprecise, which cannot be directly used as global path for navigation. It is worth noting that the proposed process is mainly for demonstrating the robustness of our model. For real-world applications, we suggest to use OSM or other tools to create the topometric map. What we want to emphasize here is that our method does not need to construct HD-maps which require huge cost to build and maintain.

\subsection{Problem setting} Let $t$ denote the discrete time step, and let $s_t = (x_t, y_t)$ denote the UTM coordinate of the autonomous vehicle at time step $t$.  Given the observed raw sensor data at time step $t$ and a coarse route extracted from the aforementioned noisy topometric map, the goal is to predict a future guidance trajectory over the next $T$ time steps, $\rvs = \{ s_t, \dots s_{t + T-1}\}$ and use this trajectory for local navigation.

\subsection{Architecture}

The overall architecture of the proposed model is a multi-task learning framework as illustrated in Fig.~\ref{fig:architecture}. The input of the model is the raw LiDAR point cloud and a local route extracted from noisy topometric map. The LiDAR point cloud is converted to a BEV (Bird's Eye View) map and the local route is converted to a binary image. The LiDAR BEV map and binary image are then concatenated and feeded into the convolutional neural network backbone to extract deep features. A waypoint position encoder takes the deep features as input and outputs waypoint heatmaps. The heatmaps are further sent to depth-wise convolution layers to obtain waypoint position embeddings. The waypoint feature encoder module takes as input the features from the backbone and outputs waypoint-wise features. The waypoint transformer follows the structure of \cite{Transformer}, which takes as input the waypoint positional embeddings and waypoint features, and predicts waypoints coordinates.

\subsubsection{Input representations}
\label{sec:input}

\begin{figure}
    \centering
    \begin{minipage}[!t]{0.4\textwidth}
        \includegraphics[width=\textwidth]{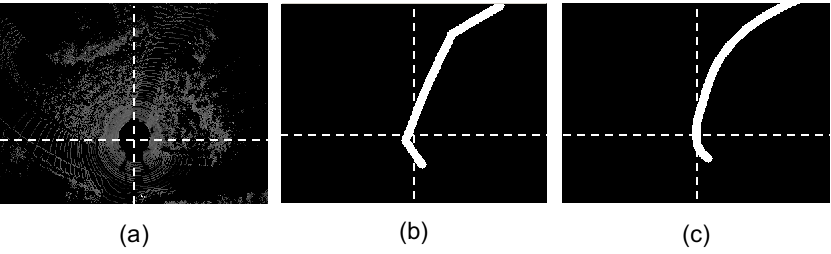}
        \caption{(a) The LiDAR BEV map. (b) The binary image of local route. (c) The binary image of recorded driving trajectory. The dash lines represent the axis of the ego-car coordinate.}
        \label{fig:input}
    \end{minipage}
\end{figure}

As our goal is to inference the local guidance trajectory in the ego-centric coordinate, it is straightforward to represent the scene from the BEV. Following \cite{2019End}, 
we convert the raw LiDAR point cloud to BEV representation. Our BEV map consists of $3$ channels, including height, intensity (reflectance) and density of point cloud. This results in a 3D tensor of size $H_0 \times W_0 \times 3$, where $H_0, W_0$ represents the y-x spatial dimension. In order to keep more context information, the BEV map also include some part of the scene behind the vehicle as illustrated in Fig.~\ref{fig:input} (a).

Given the location of the vehicle, we extract the local route from the topometric map. Because of the imprecise topometric map, we include both forward and backward waypoints in the local route. As it is shown in Fig.~\ref{fig:input} (b), the local route is further projected to the LiDAR BEV map with a fixed width of {\em virtual road} and converted to a binary image. In this work, we set default {\em virtual road} width as $2$ meters. The same process is applied to the recorded driving trajectories and we obtain binary image of the recorded driving trajectory as shown in (c). The binary image of local route and the LiDAR BEV map are used as input of the neural network for both training and testing. The binary image of the recorded driving trajectory is used as ground truth in the road attention header during training.

\subsubsection{Backbone}

Our backbone is adapted from the 3D object detection network of \cite{2019PIXOR}. It uses a feature pyramid network that combines high-resolution features with low-resolution ones. It consists of five blocks. The first block consists of two convolutional layers and the second to fifth blocks are composed of residual layers. 
The final feature map is $4 \times$ down-sampling factor with respect to the input, {\ie} $\frac{H_0}{4} \times \frac{W_0}{4} \times C$, where $C=96$ is the number of channels.

\subsubsection{Waypoint feature encoder}
The transformer module expects a sequence as input, hence we design a waypoint feature encoder to generate waypoint embeddings. The waypoint feature encoder takes as input the above deep feature map and outputs waypoint embeddings of size $N \times d$ for the transformer, where $N$ is the number of waypoints per trajectory. As illustrated in Fig.\ref{fig:architecture}, the waypoint feature encoder has two branches. The first branch consists of one convolution layer, which outputs feature map $\rvx_1$ with size $\frac{H_0}{8} \times \frac{W_0}{8} \times N$. The feature map is reshaped to size of $N \times d$. The second branch has two parts. The frontal part is a $3 \times 3$ convolution layer which outputs a road segmentation mask of size $\frac{H_0}{4} \times \frac{W_0}{4} \times 1$. The second part contains another convolution layer that takes the segmentation mask as input and output feature map $\rvx_2$ with size $\frac{H_0}{8} \times \frac{W_0}{8} \times N$. $\rvx_2$ is also reshape to $N \times d$. $\rvx_1$ and $\rvx_2$ are combined by element-wise addition to obtain the final waypoint embeddings. The road segmentation mask is learned by minimizing the binary cross entropy loss between predicted road mask and the ground truth virtual road mask as shown in Fig.~\ref{fig:input} (c).

\subsubsection{Waypoint positional encoding} Since the transformer architecture is permutation-invariant, we design a new module to generate waypoint positional embeddings to supplement the above waypoint embeddings. The waypoint position encoder module first uses a $3 \times 3$ convolution layer followed by spatial softmax layer to output waypoint heatmaps of size $\frac{H_0}{4} \times \frac{W_0}{4} \times N$. The waypoint heatmap model is learned by minimizing MSELoss (Mean Squared Error loss) between predicted heatmaps and ground truth heatmaps. We create the ground truth heatmaps using Gaussian distribution. The predicted waypoint heatmaps are sent to the second stage convolution layer to obtain positional features of size $\frac{H_0}{8} \times \frac{W_0}{8} \times N$. The positional feature are reshaped to $N \times d$ to obtain final waypoint positional embeddings, {\ie}, $\rvx_3$ in Fig.~\ref{fig:architecture}.

\subsubsection{Waypoint transformer} The waypoint transformer follows the structure of Transformer \cite{Transformer}. It consists of a transformer encoder, transformer decoder and a feed forward network (FFN). The transformer encoder consists of multiple encoder layers. Each encoder layer has a standard architecture and consists of a multi-head self-attention module and a FFN. The transformer decoder has similar structure to the transformer encoder and transforms $N$ embeddings of size $d$ using multi-headed self-attention mechanisms. Similar to DETR \cite{DETR}, our model decodes $N$ waypoints in parallel. Since the decoder is also permutation-invariant, we add a learned positional encodings {\ie}, the query embeddings, to make it output ordered waypoint embeddings. The FFN takes the embeddings and regresses the final waypoint coordinates. We use MSELoss to train the waypoint regressor.

\begin{figure}
    \centering
    \begin{minipage}[!t]{0.4\textwidth}
        \includegraphics[width=\textwidth]{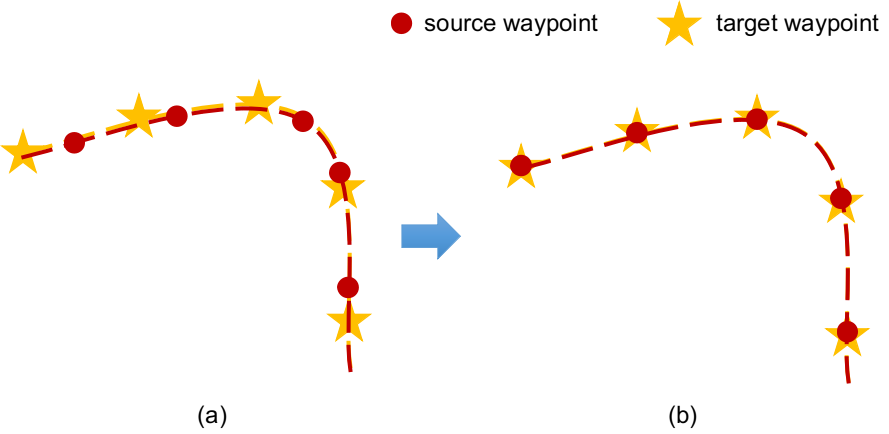}
        \caption{(a) The original predicted and ground truth waypoints. (b) The predicted waypoints after alignment.}
        \label{fig:align}
    \end{minipage}
\end{figure}

\section{Learning and inference}

We adopt the multi-task loss to train the full network. Specifically, we use the binary cross-entropy loss on the auxiliary task of road segmentation and MSELoss for the heatmap prediction and waypoint regression tasks.

Given the predicted road mask $\rvx$ of size $W' \times H'$ and binary image of the recorded driving trajectory $\rvy$ of the same size, the road segmentation loss is written as:

\begin{equation}
    \label{eq:loss_att}
    \mathcal{L}_{road} = - \sum_{i=1}^{H'} \sum_{j=1}^{W'} [\rvy_{i,j} \log{(\rvx_{i,j})} + (1 - \rvy_{i,j}) \log{(1 - \rvx_{i,j})}].
\end{equation}

Given the predicted heatmaps $\rvx$ of size $W' \times H' \times N$ and ground truth heatmaps $\rvy$ generated from driving trajectory of the same size, the heatmap prediction loss is written as:

\begin{equation}
    \label{eq:loss_heat}
    \mathcal{L}_{heatmap} = \sum_{i=1}^{H'} \sum_{j=1}^{W'} \| \rvy_{i,j} - \rvx_{i,j} \|_2^2.
\end{equation}

Assume $\rvs$ and $\rvs^*$ are the predicted and ground truth trajectory respectively, the waypoint regression loss is then defined as follows,

\begin{equation}
    \label{eq:loss_waypoint}
    \mathcal{L}_{waypoint} = \sum_{t=0}^{T-1} \| s_t^i - s_t^j\|_2^2.
\end{equation}

The final loss is the sum of the road segmentation loss, heatmap prediction loss and waypoint regression loss, {\ie} $\mathcal{L}_{all} = \mathcal{L}_{road} + \mathcal{L}_{heatmap} + \mathcal{L}_{waypoint}$.

The point-wise $L^2$ distance function in $\mathcal{L}_{waypoint}$ has a strong assumption that the driving speed on two trajectories are the same. Fig.~\ref{fig:align} (a) shows an example of two overlapping trajectories but with different driving speed. To overcome this drawback and better evaluate the prediction performance, we propose an aligned distance function by interpolating the source waypoints to the coordinates of the target waypoints as illustrated in (b). In this way, we can better measure the similarity of two trajectories. The proposed aligned distance function is used in our evaluation metrics.

The model is implemented in Pytorch and trained on $4$ RTX 2080 Ti GPUs with $11$ GB memory. We use a per-GPU batch size of $64$ and trained with Adam optimizer. The initial learning rate is $0.003$. All models are trained end-to-end from scratch for $120$ epochs. The model is deployed on IPC with a single GTX 1080 Ti GPU and the inference time is around $20$ ms, {\ie} $50$ Hz, which satisfies requirement of real-time performance.

\section{Experiments}
In this section, we evaluate the proposed trajectory prediction method on real-world autonomous driving dataset, including both urban and rural scenes.
\subsection{Experimental setup}

\subsubsection{Dataset}
\begin{figure}
    \centering
    \begin{minipage}{0.4\textwidth}
        \includegraphics[width=\textwidth]{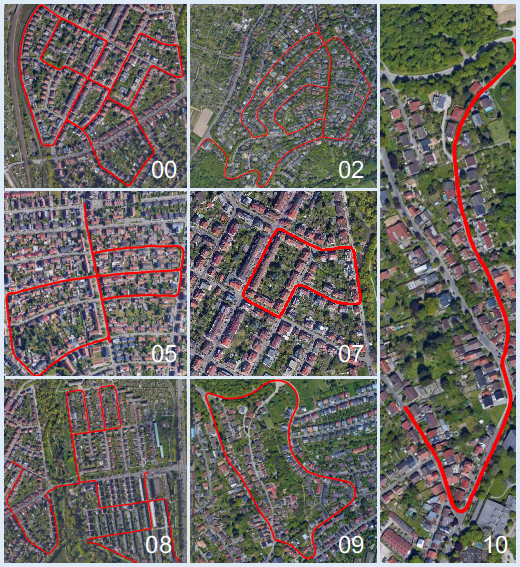}
        \caption{Recorded driving trajectories. Sequence $00$, $02$, $05$ and $07$ are used for training, and the rest are used as testing sets.}
        \label{fig:kitti}
    \end{minipage}
\end{figure}
For urban environment, we use KITTI raw data \cite{Geiger:2012}.
Following the naming rules of KITTI odometry benchmark, we use the sequence $00$, $02$, $05$ and $07$ as training data and $08$, $09$ and $10$ as testing data. There are in total $4720$ training samples and $1687$, $652$ and $349$ testing samples for $08$, $09$ and $10$ testing dataset respectively. The recorded driving trajectories are shown in Fig.~\ref{fig:kitti}.

As no rural area autonomous driving dataset is publicly available, we collect data at off-road environment by manual driving. The vehicle has a HESAI Pandar64 LiDAR and GPS/IMU system to capture point cloud and ground truth trajectories. We collected $656$ trajectories for training and another non-overlapping $701$ trajectories for testing. For both urban and rural dataset, we create the topometric maps following the proposed strategy in \secref{sec:map}. 

\subsubsection{Metrics}
In this work, we follow the common evaluation protocol in trajectory prediction literature and use \textbf{FDE}, \textbf{ADE} and \textbf{HitRate} as evaluation metrics. 

The final displacement error (\textbf{FDE}) is calculated by $\| s_{T-1} - s^*_{T-1}\|_2^2$, where $\rvs$ and $\rvs^*$ are the predicted and ground truth trajectory respectively. Because the final predicted waypoint can be used as the goal for local planning when considering local planning for obstacle avoidance, the \textbf{FDE} metric is useful for evaluating model's accuracy on predicting such goal waypoints.

The Average displacement error (\textbf{ADE}) is defined by $\frac{1}{T} \sum_{t=0}^{T-1} \| s_t - s^*_t\|_2^2$. For evaluating with multimodal trajectory predictors, we use the minimum average displacement error (\textbf{minADE}) which is defined by $minADE_k = \min_{\rvs^k} \frac{1}{T} \sum_{t=0}^{T-1}\| s^k_t - s^*_t\|$. As our model can be regarded as a special case of multimodal trajectory prediction, {\ie} with modality number of $1$, we can use \textbf{minADE} to compare with other state-of-the-art multimodal models.

For better interpreting the performance in the context of planning, we also use the \textbf{HitRate} metric proposed in \cite{2019CoverNet}. We define $Hit_{k,d}$ as $1$ for a single sample if $\min_{\rvs^k} \max_{t=0}^{T-1} \| s^k_t - s^*_t\| < d$ otherwise $0$. Then we calculate the percentage of the {\em hit} samples and refer to it as the $HitRate_{k,d}$. 

\subsection{Empirical results}
\begin{table}
    \small
    \centering
\begin{tabular}{ccccc}
    \toprule
    Model    & FDE$\downarrow$  & \ade{1} $\downarrow$ & \hitrate{1} $\uparrow$ \\
    \midrule
    Heatmap Only       &   1.63  & 1.23  & 0.83    \\
    \midrule
    Transformer0   &   0.79  & 0.34   & 0.92   \\
    Transformer1   &   0.72  & 0.35   & 0.95  \\
    Transformer2   &   0.66  & 0.28   & 0.95 \\   
    Transformer3   &   0.61  & 0.26   & 0.96 \\     
    \bottomrule
\end{tabular}
\caption{Ablation studies on KITTI-10.}
\label{tab:ablation}
\end{table}

\begin{table}
    \small
    \centering
    \begin{tabular}{cccc}
    \toprule
    Perturbation    & FDE$\downarrow$  & \ade{1} $\downarrow$ & \hitrate{1} $\uparrow$\\
    \midrule
    0.0 m       &   0.61            & 0.25                  & 0.97  \\
    \midrule
    1.0 m   &   0.62$\pm$0.01       & 0.27$\pm$0.01         & 0.95$\pm$0.01  \\
    2.0 m  &   0.77$\pm$0.03       & 0.34$\pm$0.02         & 0.92$\pm$0.01  \\
    3.0 m  &   1.05$\pm$0.01       & 0.45$\pm$0.03         & 0.85$\pm$0.02  \\
    \bottomrule
    \end{tabular}
    \caption{Robustness of the model under different lateral perturbations to the topometric map.}
    \label{tab:perturbation}
\end{table}

\begin{table*}
    \small
    \centering
    \begin{tabular}{cllcccccc}
    \toprule
    Dataset                         &Method    & FDE$\downarrow$  & \ade{1} $\downarrow$ & \ade{2} $\downarrow$ & \ade{3} $\downarrow$ & \hitrate{1} $\uparrow$ & \hitrate{2} $\uparrow$ & \hitrate{3} $\uparrow$ \\
    \midrule
    \multirow{4}{*}{KITTI-08}  & Const. vel. \& yaw &   2.40           & 1.04                 & 1.04                 & 0.1.04                 & 0.71                   & 0.71                 & 0.71  \\    
    & CoverNet \cite{2019CoverNet} &   1.28           & 0.62                 & 0.47                 & {0.41}                 & 0.80                   & 0.84                 & 0.87  \\
    & MTP \cite{2018Multimodal}    &   1.06           & 0.53                 & 0.45                 & 0.42                 & {0.85}                   & \textbf{0.88}                 & \textbf{0.89}  \\
    & MultiPath \cite{2019MultiPath}   &   {1.00}    & {0.51}           & {0.44}                    & {0.41}                 & {0.85}                   & 0.86                & \textbf{0.89}  \\
    &                          Ours    &   \textbf{0.81}  & \textbf{0.39}    & \textbf{0.39}        & \textbf{0.39}          & \textbf{0.87}            & 0.87                & 0.87  \\

    \midrule
    \multirow{4}{*}{KITTI-09}   & Const. vel. \& yaw &   2.66           & 1.00                 & 1.00                 & 1.00                 & 0.55                   & 0.55                 & 0.55  \\   
    & CoverNet \cite{2019CoverNet} &   1.42           & 0.56                 & 0.38                 & 0.33                 & 0.71                   & 0.86                 & 0.91  \\
    & MTP \cite{2018Multimodal}    &   0.95           & 0.39                 & 0.29                 & 0.25                 & 0.87                   & \textbf{0.94}                 & \textbf{0.96}  \\
    & MultiPath \cite{2019MultiPath}   &   {0.85}     & {0.35}           & \textbf{0.27}          & \textbf{0.23}    & {0.90}                   & \textbf{0.94}                 & \textbf{0.96}  \\
    &                          Ours    &   \textbf{0.71}  & \textbf{0.29}    & 0.29                 & 0.29            & \textbf{0.92}           & 0.92                & 0.92  \\

    \midrule
    \multirow{4}{*}{KITTI-10} & Const. vel. \& yaw &   2.11           & 0.90                 & 0.90                 & 0.90                 & 0.64                   & 0.64                 & 0.64  \\
    & CoverNet \cite{2019CoverNet} &   1.36           & 0.53                 & 0.37                 & 0.30                 & 0.78                   & 0.89                 & 0.94  \\
    & MTP \cite{2018Multimodal}    &   0.73           & 0.31                 & 0.26                 & \textbf{0.23}    & 0.93                   & \textbf{0.97}                 & \textbf{0.97}  \\
    & MultiPath \cite{2019MultiPath}   &   {0.64}     & {0.30}        & \textbf{0.24}        & \textbf{0.23}         & {0.95}                   & \textbf{0.97}                 & \textbf{0.97}  \\
    &                          Ours &   \textbf{0.61} & \textbf{0.25}        & 0.25          & 0.25                  & \textbf{0.97}            & \textbf{0.97}                & \textbf{0.97}  \\

    \midrule
    \multirow{4}{*}{Rural}   & Const. vel. \& yaw &   2.90           & 1.37                 & 1.37                 & 1.37                 & 0.60                   & 0.60                 & 0.60  \\   
    & CoverNet \cite{2019CoverNet} &   2.69      & 1.37                 & 0.96                 & 0.78                 & 0.65                   & 0.75                & 0.79  \\
    & MTP \cite{2018Multimodal}    &   1.61           & {0.84}          & {0.71}            & \textbf{0.66}        & {0.76}          & {0.82}       & {0.82}  \\
    & MultiPath \cite{2019MultiPath}   &   {1.59}     & 0.90             & 0.75                 & 0.71                 & 0.75                   & 0.81                & {0.82}  \\
    &                          Ours    &   \textbf{1.19}  & \textbf{0.69}  & \textbf{0.69}   & 0.69                 & \textbf{0.85}       & \textbf{0.85}           & \textbf{0.85}  \\
    \bottomrule
    \end{tabular}
    \caption{Results on KITTI and our collected rural scene dataset. Smaller minADE and FDE is better. Larger HitRate is better.}
    \label{tab:results}
\end{table*}

We implemented several state-of-the-art multimodal baselines using the same input and backbone architectures, including CoverNet \cite{2019CoverNet}, MTP \cite{2018Multimodal} and MultiPath \cite{2019MultiPath}. As multimodal approaches predict multiple trajectories, we use a fixed number of $3$ in all our experiments. The main results are summarized in Table~\ref{tab:results}. The model denoted by {\em Const. vel. \& yaw} is a classic physics-based model. we use the vehicle's position observed in the past trajectory as measurement to estimate the vehicle's velocity and yaw angle. During the inference stage, we assume the velocity and yaw are unchanged. 

From the overall results, we can see that physics-based single modality models are clearly not suitable for long-term predictions. In general, the models with regression-based header (MTP and MultiPath) significantly outperform the one with classification-based header {\ie} the CoverNet baseline. 
All these models obtain higher accuracies on urban environment than on rural environment. This is reasonable, because urban scene is more structured {\eg} the boundaries of roads are clear and roads are mostly flat. We can also see from the results on KITTI that the simpler the scenario the higher the accuracy. KITTI-08 has the most complex trajectories, and the accuracy is the worst,  while KITTI-10 is on the opposite side. Lastly, the model with anchors generally outperforms the one without anchors, {\eg}, MultiPath vs MTP, especially for urban scenes. Comparing to the baseline models, our transformer based model constantly obtains higher accuracies.

\begin{figure*}
    \centering
    \begin{minipage}{0.9\textwidth}
        \includegraphics[width=\textwidth]{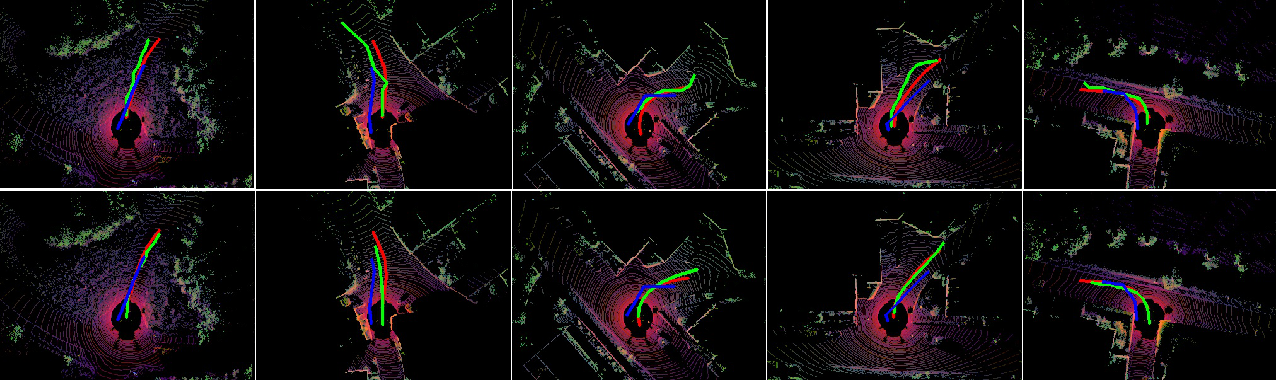}
        \caption{Visualization of the predicted trajectories. The top row shows the results of the model without using transformer. The bottom row shows the results of the model with transformer. Blue: local route. Red: ground truth trajectory. Green: predicted trajectory.}
        \label{fig:results}
    \end{minipage}
\end{figure*}

\subsection{Ablation study}

In this section, we first conduct ablation studies to analyze different components in our proposed model. A baseline model without using transformer is denoted by {\em Heatmap only} in Table~\ref{tab:ablation}. We designed several variants of our transformer based model. {\em Transformer0} is the model without using road segmentation and without using waypoint positional encoder. {\em Transformer1} is the model without using road segmentation. {\em Transformer2} is the model without using waypoint positional encoder. {\em Transformer3} is the model without using transformer decoder. The {\em Heatmap only} model performs worst due to simple model architecture and ignoring the dependencies between waypoints. Removing the positional encoder (Transformer2) causes $0.05$ FDE drop and removing the road segmentation (Transformer1) causes $0.11$ FDE. Removing both (Transformer0) causes $0.18$ FDE drop. While removing the transformer decoder causes least performance drop.

We also show qualitative results in Fig.~\ref{fig:results}. The first row shows the prediction results from {\em Heatmap only} and the second row shows our transformer based model. The transformer based model predicts more accurate and smoother trajectories than the non-transformer one, showing the effectiveness of the proposed waypoint transformer architecture.

Next, we analyze the robustness of the model to the noise of the topometric map. We add random lateral perturbation to the topometric map in the testing set and run testing with the perturbed local route. Table~\ref{tab:perturbation} shows the accuracies of the model under different magnitude of lateral perturbation. Each random experiment is repeated for $3$ times and we report the mean and deviation. We have tried to repeat for $3, 5$ and $10$ times, and the statistics does not show much differences. Note that for other experiments in this work, we use fixed topometric maps, so they do not need to compute the standard deviations. For lateral perturbation of $1.0$ meters, the model keeps almost the same accuracies with deviation around $0.01$. When increasing the lateral perturbation to $3.0$ meter, FDE drops around $0.4$ meter, but still far less than the magnitude of the perturbation, {\eg} $0.4$ vs $3.0$. It worth noting that the magnitude of random perturbation during training is $0.25$ meter. From this study, we can see the proposed model is robust the noise of the topometric map.

\section{Conclusions}
In this work, we have proposed an end-to-end transformer based deep model that takes raw LiDAR and noise topometric map as input and predicts future guidance trajectory. The proposed model consists of a novel waypoint position encoder, waypoint feature encoder and transformer. We evaluated the proposed model on both urban (KITTI) and self-collected rural scene datasets and obtained promising accuracies. Our current approach can not handle dynamic obstacles on the road, such as pedestrians and other vehicles. As future work, we are considering to address this issue by incorporating perception tasks {\eg} object detection and semantic segmentation in the multi-task learning framework, {\ie} to implement an end-to-end neural motion planner.

\bibliographystyle{IEEEtran}
\bibliography{references}

\end{document}